\documentclass{article} 
\usepackage{colm2024_conference}
\usepackage{microtype}
\usepackage{hyperref}
\usepackage{url}
\usepackage{graphicx} 
\usepackage{amsfonts}
\usepackage{amsmath}
\usepackage{booktabs}
\usepackage{float}
\usepackage{xcolor}
\usepackage[utf8]{inputenc}
\usepackage{multicol}

\title{Hy ASR 3.0 Technical Report}
\author{
	\bf \large Hy ASR team
}
\date{}

\begin{document}

\maketitle

\begin{abstract}
We present Hy ASR 3.0, a large language model (LLM)-based automatic speech recognition (ASR) system that delivers strong recognition accuracy, supports flexible contextual conditioning, and maintains robust recognition performance across diverse speech conditions. Hy ASR 3.0 supports Mandarin Chinese, English and 13 Chinese dialect groups, including Cantonese. Among the evaluated systems, it achieves the lowest aggregate error rates on public Mandarin Chinese, English, and Cantonese benchmarks and the lowest error rates on the corresponding internal test sets, alongside first-tier dialect recognition performance. Beyond audio-only transcription, Hy ASR 3.0 supports multiple forms of external context, including hotword lists, question--answer dialogues, and summaries, enabling the use of task-relevant information during recognition. The model also leverages intra-utterance context to disambiguate homophonic and phonetically similar keywords using information from other parts of the same utterance, without requiring a separate contextual input. Evaluations demonstrate strong contextual recognition in both external-context and intra-utterance settings, as well as robust recognition of long-form speech, children's and older adults' speech, low-volume speech, and speech in noisy environments. This report describes the model architecture and training methodology and presents evaluations across recognition accuracy, context utilization, and robustness.
\end{abstract}

\begin{figure}[H]
    \centering
    \includegraphics[width=\linewidth]{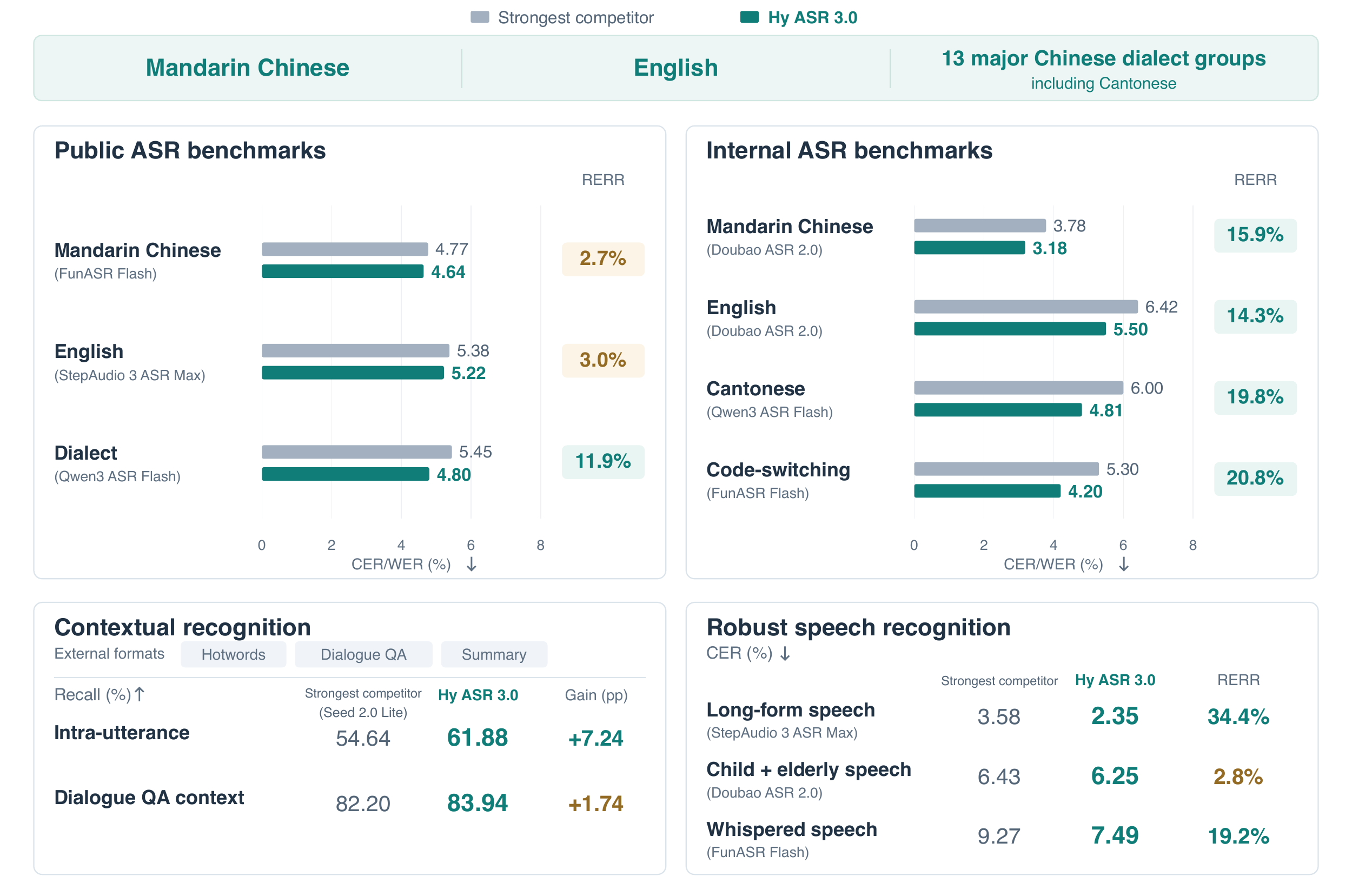}
    \caption{Overview of Hy ASR 3.0 capabilities and benchmark performance.
    Comparisons are made against the strongest competitor for each metric,
    with the corresponding model shown in parentheses. RERR denotes the relative error rate reduction.}
    \label{fig:hy-asr-overview}
\end{figure}
\label{sec:model-architecture}

\section{Introduction}
Automatic speech recognition (ASR) has advanced considerably through large-scale training and the integration of pretrained language models. Whisper demonstrated the effectiveness of large-scale weak supervision for recognition accuracy and cross-domain generalization~\citep{radford2023whisper}. LLM-based approaches have further enabled transcription to incorporate pretrained linguistic knowledge and textual context~\citep{lakomkin2023contextualization}, while recent systems such as Qwen3 ASR have broadened recognition coverage and evaluation across diverse speech conditions~\citep{qwen2026asr}. Building on these developments, we focus on three practical objectives: improving recognition accuracy across Mandarin Chinese, English, Cantonese, and regional dialects; leveraging heterogeneous contextual inputs effectively; and enhancing transcription robustness across speaker age groups, speaking styles, and audio durations.

In this report, we introduce Hy ASR 3.0, an LLM-based ASR system developed around these objectives. We evaluate the system on public benchmarks and internal test sets covering standard speech recognition, dialect recognition, context-aware recognition, and robustness under challenging speech conditions. Its principal capabilities are summarized as follows:

The key capabilities and advances of Hy ASR 3.0 are summarized
as follows:

\begin{itemize}
\item \textbf{Leading Mandarin, English and Cantonese recognition, with competitive dialect performance.} Hy ASR 3.0 achieves the lowest aggregate error rates among the evaluated systems on the public Mandarin Chinese, English and Cantonese benchmarks, reaching 4.64\%, 5.22\%, and 5.06\%, respectively. These correspond to approximately 2.7\%, 3.0\%, and 11.0\% relative error rate reductions (RERR) over the strongest non-Hy competitor in each setting: FunASR Flash, StepAudio 3 ASR Max~\citep{lin2026stepaudio3realtime}, and Qwen3 ASR Flash~\citep{qwencloud_qwen3_asr_flash}, respectively. The Cantonese aggregate excludes KeSpeech, unlike the combined Cantonese-and-KeSpeech group in Table~\ref{tab:open-benchmarks}. Beyond Mandarin Chinese, English, and Cantonese, the system supports 12 other major Chinese dialect groups. Internal human evaluation covers these dialect categories, spanning 91 subregional varieties, demonstrating state-of-the-art (SOTA) performance. On KeSpeech~\citep{tang2021kespeech}, Hy ASR 3.0 achieves an error rate of 3.77\%, compared with 4.33\% for FunASR Flash.

\item \textbf{Strong contextual recognition through external conditioning and intra-utterance disambiguation.} Hy ASR 3.0 supports external hotword lists, question--answer dialogues, and summaries. Hotword lists provide explicit vocabulary, while dialogues and summaries provide broader semantic information. The model also uses information within an utterance to disambiguate homophonic and phonetically similar keywords, without treating this information as a separate input. It achieves the highest target-keyword recall among the evaluated systems on both the intra-utterance context and dialogue-context test sets (Table~\ref{tab:context}).

\item \textbf{Robust recognition across audio durations, speaker age groups, and speaking styles.} Hy ASR 3.0 achieves the lowest character error rate (CER) among the evaluated systems on internal benchmarks covering long-form speech, a combined test set of children's and older adults' speech, and a combined test set of low-volume speech and speech in noisy environments.
(Table~\ref{tab:robustness}). Relative to the strongest non-Hy competitor in each setting, it reduces CER by 34.4\%, 2.8\%, and 19.2\%, respectively, with particularly large gains on long-form and the combined evaluation of low-volume speech and speech in noisy environments.
\end{itemize}

Together, these evaluations demonstrate Hy ASR 3.0’s strengths in recognition accuracy, contextual adaptability, and robustness. The remainder of this report describes the model architecture and training methodology, followed by detailed experimental results across these three capability areas.

\section{Model Architecture}

\begin{figure}[H]
    \centering
    \includegraphics[width=\linewidth]{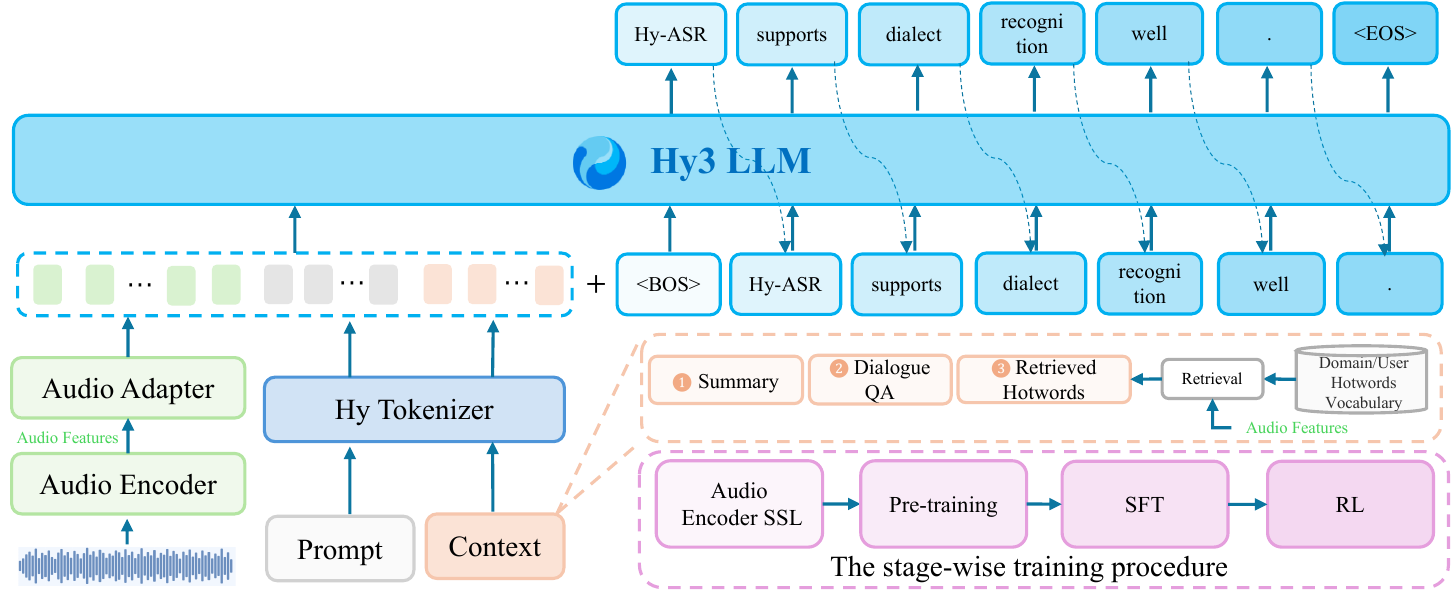}
    \caption{Overview of the Hy ASR 3.0 model architecture and its training procedure.}
    \label{fig:architecture}
\end{figure}

As illustrated in Figure~\ref{fig:architecture}, Hy ASR 3.0
comprises an audio encoder, an audio adapter, and an LLM decoder,
together with an audio-conditioned hotword retrieval mechanism.
The system integrates speech representations and external textual
context to generate the output transcription.

\paragraph{Audio representations.}
The audio encoder extracts a sequence of acoustic representations
from the input speech. The audio adapter maps these representations
into embeddings compatible with the LLM input space. The encoder
representations are also provided to the hotword retrieval mechanism
for selecting relevant vocabulary.

\paragraph{Contextual inputs and hotword retrieval.}
Hy ASR 3.0 supports external context in the form of hotwords,
summaries, and question--answer (QA) dialogues. Given the
audio representations and a candidate hotword list, the retrieval
mechanism selects relevant terms to form the hotword context.
These terms, together with any available summary or dialogue QA
context, are encoded as a textual prefix placed before the audio
embeddings.

\paragraph{Transcription generation.}
The LLM decoder generates the transcription autoregressively,
conditioned on the context prefix, audio embeddings, and previously
generated tokens. This input organization allows external
information to guide recognition while preserving direct access
to the speech representations. The internal context to the utterance
is captured through the audio and transcription sequence, rather
than supplied as a separate external input.
\label{sec:model-architecture}

\section{Training}
\label{sec:training}

We build our ASR model upon the pretrained Hy3 base model~\citep{tencent_hy3}. The
training pipeline consists of three stages: audio encoder
pre-training, large-scale ASR pre-training, and post-training.

\subsection{Encoder}
\label{sec:encoder_pretraining}
The audio encoder is trained to extract semantic speech
representations for the downstream language model. We adopt an
iterative HuBERT-style self-supervised learning pipeline~\citep{hsu2021hubert}. We first
train a bootstrap encoder with Best-RQ~\citep{chiu2022bestrq} and quantize its selected
intermediate representations to construct discrete targets. These
targets are then used to train an intermediate HuBERT model, whose
representations are further quantized with residual vector
quantization to produce multi-level targets. The final audio encoder
is trained to predict these targets from masked speech inputs.

The audio encoder is trained in two phases on tens of millions of
hours of speech, primarily in Mandarin and English, with Cantonese and
other Chinese dialects also included. We first perform large-scale
pre-training on the broad speech mixture, followed by continued
pre-training on a rebalanced mixture that places greater emphasis on
Chinese dialects while down-weighting noisy recordings and short
utterances. We further strengthen the masking strategy and apply
utterance mixing \citep{Chen_2022} to improve robustness to interfering speech.
\subsection{Pre-training}
\label{sec:pretraining}
We adopt a two-stage pre-training strategy, comprising large-scale
pre-training followed by targeted continued pre-training.
\paragraph{Stage 1: Large-scale pre-training.}
In the first stage, we pre-train the model on a large-scale
web-sourced speech corpus comprising tens of millions of hours of
speech data. To construct reliable supervision at scale,
we develop a multi-system ASR data construction pipeline, in which
multiple ASR systems independently generate candidate transcription
hypotheses. These hypotheses are combined using Recognizer Output Voting Error Reduction (ROVER)~\citep{fiscus1997rover}, and the resulting pseudo-transcriptions are subsequently
subjected to multi-stage quality filtering to discard unreliable
samples. We further apply domain-aware data balancing and data compression to improve training-data efficiency. Data balancing helps allocate training capacity more effectively across heterogeneous domains, while data compression removes redundant information and prioritizes more informative samples. Together, these strategies substantially improve data efficiency: the resulting model achieves better performance while using only 50\% of the original training data, demonstrating that carefully curated and balanced data can be more effective than simply scaling up the training corpus.

\paragraph{Stage 2: Targeted continued pre-training.}
In the second stage, we continue pre-training on a rebalanced data
mixture with increased proportions of dialectal speech,
context-biasing data, and speech recorded under challenging acoustic
conditions. This targeted data composition improves dialect
recognition, strengthens the model's context-biasing capability, and
enhances robustness to adverse acoustic conditions. Moreover, by aligning the pre-training distribution more closely with the data used in subsequent supervised fine-tuning (SFT), this stage improves the effectiveness of SFT and enables us to reduce the SFT training data to just one-tenth of its original size while maintaining  model performance. This reduction substantially accelerates SFT experimentation and iteration, allowing us to evaluate training strategies and incorporate improvements more efficiently.

\subsection{Post-training}
\label{sec:posttraining}
Post-training is performed in two stages, including supervised fine-tuning  and reinforcement learning.

\paragraph{Stage 1: Supervised Fine-tuning (SFT).}
During SFT, we improve recognition performance on dialectal speech,
contextual-biasing tasks, and speech under challenging acoustic
conditions, while maintaining the general-domain ASR performance
obtained during pre-training. To obtain reliable supervision, we
continuously improve the accuracy of the individual labeling components
in our automatic transcription pipeline. At each iteration, the
best-performing SFT checkpoint from the previous iteration is used as
the teacher ASR model to generate pseudo-transcriptions for the next
iteration. This iterative pseudo-labeling procedure progressively
improves the quality of the SFT data. We further supplement the
training mixture with targeted data for dialectal speech and other
challenging recognition scenarios.


\paragraph{Stage 2: Reinforcement Learning (RL).}
\textbf{a) Expert RL} 
We train three independent ASR experts initialized from the SFT
checkpoint, targeting general-domain ASR, dialectal ASR, and
context-biasing ASR, respectively. Training the experts separately
reduces cross-domain interference and allows each model to specialize
in its target scenario. The reward system of each expert is designed
according to the characteristics of the corresponding domain. Besides
overall recognition accuracy, the rewards also consider keyword
recognition and the utterance-level fluency of the transcription.

\textbf{b) Multi-teacher on-policy distillation.}
We then integrate the capabilities of the three experts into a single
model through multi-teacher on-policy distillation (MOPD)~\citep{agarwal2024gkd}. For each
utterance, the student first generates an on-policy recognition
hypothesis, which is then routed to the corresponding domain expert
for token-level supervision. The objective is written as
\begin{equation}
\mathcal{L}_{\mathrm{MOPD}}
=
\mathbb{E}
\left[
\frac{1}{T}
\sum_{t=1}^{T}
D_{\mathrm{KL}}
\left(
\pi_{\theta}(\cdot \mid \mathbf{s}_t)
\Vert
\pi_{\phi_d}(\cdot \mid \mathbf{s}_t)
\right)
\right].
\label{eq:mopd}
\end{equation}
Here, $T=|\hat{\mathbf{y}}|$ and
$\mathbf{s}_t=(\mathbf{x},\hat{\mathbf{y}}_{<t})$,
where $\pi_{\theta}$ is the student, $\pi_{\phi_d}$ is the expert
associated with domain $d$, and $\mathbf{x}$ denotes the acoustic input
together with optional biasing context. The expectation is taken over
domain-routed training samples and hypotheses generated by the student.
This stage transfers the domain-specific strengths of the three
experts into a unified ASR model. 
Joint reinforcement learning across heterogeneous ASR objectives requires a shared policy to optimize general recognition accuracy, dialect-specific behavior, and context-dependent keyword recognition simultaneously. These objectives may impose different optimization preferences, particularly when their data distributions and reward signals differ. Our expert-based strategy separates domain-specific optimization from capability integration: independently trained experts can specialize in their respective recognition scenarios, while multi-teacher on-policy distillation transfers their behaviors into a unified student. This design provides a structured alternative to directly mixing all RL objectives in a single policy.

\textbf{c) Model-Behavior RL.}
Finally, we apply Model-Behavior RL to hard cases identified from
real-world usage, with a particular focus on keywords whose correct
recognition depends on the surrounding context. By explicitly
reinforcing context-dependent keyword recognition, this stage improves
the model's use of contextual information and produces more coherent
and fluent utterance-level transcriptions.

\section{Evaluation}
\paragraph{Compared systems and metrics.}
We compare Hy ASR 3.0 with Qwen3 ASR Flash~\citep{qwencloud_qwen3_asr_flash} and FunASR Flash~\citep{alibaba_cloud_qwen3_asr_flash_funasr_flash_guide}, Doubao ASR 2.0~\citep{volcengine_doubao_asr_2_product}, Seed 2.0 Lite~\citep{bytedance_seed_2_0_overview}, StepAudio 3 ASR Max~\citep{lin2026stepaudio3realtime,stepfun_stepaudio_3_asr_max_overview,stepfun_stepaudio_3_asr_max_sse_api}, and Hy ASR 3.0 Preview. These technical reports and official service references identify the model families or services; the numerical comparisons below are the results reported in this study, not scores taken from those references. We report character or word error rates (CER/WER) for objective recognition evaluation, target-keyword recall for contextual recognition, and human-rated percentages for dialect evaluation. Relative error rate reduction is computed as $(E_{\mathrm{comp}}-E_{\mathrm{Hy}})/E_{\mathrm{comp}}$, where $E_{\mathrm{comp}}$ is the error rate of the strongest non-Hy competitor for the metric being compared. Differences between recall or human-rated percentages are expressed in percentage points.

\subsection{Public ASR Benchmarks}
Our public evaluation includes AISHELL-1~\citep{bu2017aishell1}, AISHELL-2~\citep{du2018aishell2}, AliMeeting~\citep{yu2022m2met}, WenetSpeech~\citep{zhang2022wenetspeech}, FLEURS~\citep{conneau2023fleurs}, Common Voice~\citep{ardila2020commonvoice}, SpeechIO~\citep{speechio_leaderboard}, LibriSpeech~\citep{panayotov2015librispeech}, GigaSpeech~\citep{chen2021gigaspeech}, TED-LIUM 3~\citep{hernandez2018tedlium3}, VoxPopuli~\citep{wang2021voxpopuli}, MLS~\citep{pratap2020mls}, WenetSpeech-Yue~\citep{li2026wenetspeechyue}, and KeSpeech~\citep{tang2021kespeech}. Note that we apply a strict data decontamination procedure to ensure that no evaluation utterances, or their audio or textual near-duplicates, are included in the training data.

Table~\ref{tab:open-benchmarks} reports the results. 
Hy ASR 3.0 achieves the lowest reported aggregate error rates
across all three evaluation groups: 4.64\% for Chinese,
5.22\% for English, and 4.80\% for dialect recognition.
Relative to the strongest external baseline in each group,
these correspond to approximately 2.7\%, 3.0\%, and 11.9\%
error rate reductions, respectively. Compared with the Preview
version, the corresponding reductions are approximately
1.9\%, 1.1\%, and 8.2\%, with the largest improvement observed
in dialect recognition.

\paragraph{Mandarin Chinese.}
Hy ASR 3.0 achieves a reported aggregate error rate of 4.64\%, compared with 4.77\% for FunASR Flash and 4.73\% for Preview. It obtains the best results on the AISHELL-2 mix subset, AliMeeting near-field subset, and the subset labeled SpeechIO Unlocked in Table~\ref{tab:open-benchmarks}, with error rates of 1.98\%, 5.47\%, and 2.36\%, respectively. It also matches the best results on WenetSpeech test-net (3.74\%) and FLEURS-zh (1.84\%). These aggregate improvements are not uniform across subsets: for example, FunASR Flash remains stronger on AISHELL-1 and WenetSpeech test-meeting.

\paragraph{English.}
Hy ASR 3.0 achieves an aggregate WER of 5.22\%, improving over 5.38\% for StepAudio 3 ASR Max and 5.28\% for Preview. It obtains the lowest WER on Common Voice-en (5.48\%), FLEURS-en (2.53\%), TED-LIUM 3 (2.44\%), and VoxPopuli-en test (5.84\%), while matching Preview on GigaSpeech (8.27\%). StepAudio 3 ASR Max remains stronger on both LibriSpeech subsets and the accented VoxPopuli-en subset. On the latter, Hy ASR 3.0 also regresses from 14.75\% to 15.27\% relative to Preview, highlighting accented speech as an area for further improvement.

\paragraph{Cantonese and dialect recognition.}
The combined group comprises four Cantonese subsets and KeSpeech, which covers standard Mandarin and its eight subdialects~\citep{tang2021kespeech}. Hy ASR 3.0 reduces the reported combined aggregate to 4.80\%, compared with 5.45\% for Qwen3 ASR Flash and 5.23\% for Preview. For Cantonese alone, the unweighted mean over FLEURS-yue, Common Voice-yue, and the WenetSpeech-Yue short and long subsets is 5.06\%, compared with 5.68\% for Qwen3 ASR Flash, corresponding to approximately 11.0\% relative error rate reduction computed before rounding the aggregate values. This four-subset aggregate is distinct from the combined value in the table. On KeSpeech, the error rate decreases from Preview's 5.17\% to 3.77\%, a relative reduction of 27.1\%. Hy ASR 3.0 also achieves the best results on FLEURS-yue (2.63\%) and the WenetSpeech-Yue short subset (5.00\%). Preview remains slightly better on Common Voice-yue, while Qwen3 ASR Flash retains the lowest error rate on the WenetSpeech-Yue long subset.

\begin{table}[H]
    \centering
    \small
    \caption{ASR Performance (CER/WER, \%) on Public Benchmarks. 
    \textbf{Bold} and \underline{underlined} values mark the best and second-best results in each row, respectively.}
    \label{tab:open-benchmarks}
    \setlength{\tabcolsep}{3pt}
    \resizebox{\textwidth}{!}{%
    \renewcommand{\arraystretch}{1.23}
    \begin{tabular}{lccccccc}
        \toprule
        \textbf{Testset} & \textbf{Qwen3 ASR Flash} & \textbf{FunASR Flash} & \textbf{Doubao ASR 2.0} & \textbf{Seed 2.0 Lite} & \textbf{StepAudio 3 ASR Max} & \textbf{Hy ASR 3.0 Preview} & \textbf{Hy ASR 3.0} \\
        \midrule
        \textbf{Chinese (zh)} & 5.73 & 4.77 & 5.55 & 5.42 & 6.98 & \underline{4.73} & \textbf{4.64} \\
        \midrule
        \quad AISHELL1 & 2.12 & \textbf{1.01} & 1.90 & 2.03 & 1.40 & \underline{1.21} & 1.37 \\
        \quad AISHELL2 android/ios/mix & 2.70/2.59/2.55 & 2.16/\textbf{1.91}/2.14 & 2.67/2.55/2.69 & 2.68/2.75/2.78 & 2.30/2.20/2.35 & \textbf{2.02}/2.03/\underline{2.10} & \underline{2.05}/\underline{2.02}/\textbf{1.98} \\
        \quad AliMeeting near/far & 7.13/23.14 & 5.94/22.02 & 5.88/26.18 & 6.27/24.81 & 7.89/34.23 & \underline{5.84}/\textbf{21.27} & \textbf{5.47}/\underline{21.52} \\
        \quad WenetSpeech net/meeting & 4.96/6.16 & \textbf{3.74}/\textbf{3.62} & 4.11/4.54 & 4.13/4.23 & 6.63/8.52 & \underline{3.91}/4.31 & \textbf{3.74}/\underline{4.16} \\
        \quad FLEURS-zh & 2.95 & 2.00 & 2.40 & 2.13 & \underline{1.85} & \textbf{1.84} & \textbf{1.84} \\
        \quad CommonVoice-zh & \underline{4.41} & \textbf{4.10} & 5.61 & 4.54 & 5.00 & 4.52 & 4.49 \\
        \quad SpeechIO Locked/Unlocked & \underline{3.62}/6.47 & 5.95/\underline{2.61} & \textbf{2.52}/5.52 & 5.09/2.62 & 8.50/2.94 & 5.04/2.63 & 4.63/\textbf{2.36} \\
        \midrule
        \textbf{English (en)} & 6.21 & 6.02 & 7.93 & 5.56 & 5.38 & \underline{5.28} & \textbf{5.22} \\
        \midrule
        \quad LibriSpeech clean/other & 1.33/2.60 & 1.31/2.50 & 2.87/5.85 & 1.37/2.65 & \textbf{1.17}/\textbf{2.25} & 1.28/2.59 & \underline{1.19}/\underline{2.31} \\
        \quad CommonVoice-en & 6.32 & 6.31 & 14.18 & 6.81 & 7.21 & \underline{5.53} & \textbf{5.48} \\
        \quad FLEURS-en & 2.66 & 3.24 & 6.38 & \underline{2.55} & 2.95 & 2.64 & \textbf{2.53} \\
        \quad GigaSpeech test & \underline{9.15} & 9.96 & 9.58 & 9.43 & 9.68 & \textbf{8.27} & \textbf{8.27} \\
        \quad TED-LIUM 3 test & 3.00 & 3.39 & 3.18 & 3.04 & \underline{2.48} & 2.50 & \textbf{2.44} \\
        \quad VoxPopuli-en test/accented & 5.99/20.97 & 6.06/16.84 & 9.52/\underline{12.57} & 7.40/13.28 & 7.08/\textbf{11.75} & \underline{5.98}/14.75 & \textbf{5.84}/15.27 \\
        \quad MLS test-en & 3.87 & 4.57 & 7.27 & \textbf{3.54} & 3.85 & 3.99 & \underline{3.69} \\
        \midrule
        \textbf{Dialect} & 5.45 & 5.46 & 7.66 & 6.73 & 6.26 & \underline{5.23} & \textbf{4.80} \\
        \midrule
        \quad FLEURS-yue & 2.97 & 3.24 & 6.52 & \underline{2.64} & 3.05 & 2.76 & \textbf{2.63} \\
        \quad CommonVoice-yue & 4.67 & 4.98 & 5.84 & 6.28 & 5.90 & \textbf{3.49} & \underline{3.56} \\
        \quad WenetSpeech yue short/long & 6.51/ \textbf{8.58} & 5.73/\underline{9.01} & 9.48/11.33 & 8.27/10.77 & 7.09/9.86 & \underline{5.38}/9.34 & \textbf{5.00}/9.05 \\
        \quad KeSpeech & 4.51 & \underline{4.33} & 5.12 & 5.68 & 5.39 & 5.17 & \textbf{3.77} \\
        \bottomrule
    \end{tabular}}
\end{table}

\subsection{Internal ASR Benchmarks}
\paragraph{Main-language recognition.}
Table~\ref{tab:main_languages_results} reports objective recognition error rates on internal Mandarin Chinese, English, Cantonese, and code-switching benchmarks constructed from real-world business data. These benchmarks complement public datasets by reflecting speech conditions encountered in practical applications. Hy ASR 3.0 achieves the lowest error rates across all four categories, reaching 3.18\%, 5.50\%, 4.81\%, and 4.20\%, respectively, and improves consistently over Hy ASR 3.0 Preview.

\begin{table}[H]
    \centering
    \caption{ASR Performance (CER/WER, \%) on Internal Benchmarks. \textbf{Bold} and \underline{underlined} values mark the best and second-best results in each row, respectively.}
    \label{tab:main_languages_results}
    \small
    \setlength{\tabcolsep}{3pt}
    \resizebox{\textwidth}{!}{%
    \renewcommand{\arraystretch}{1.23}
    \begin{tabular}{lccccccc}
        \toprule
        \textbf{TestSet} & \textbf{Qwen3 ASR Flash} & \textbf{FunASR Flash} & \textbf{Doubao ASR 2.0} & \textbf{Seed 2.0 Lite} & \textbf{StepAudio 3 ASR Max} & \textbf{Hy ASR 3.0 Preview} & \textbf{Hy ASR 3.0} \\
        \midrule
        Chinese        & 4.94 & 4.15 & 3.78 & 5.56 & 5.30 & \underline{3.35} & \textbf{3.18} \\
        English        & 8.76 & 7.54 & 6.42 & 8.78 & 10.68 & \underline{5.78} & \textbf{5.50} \\
        Cantonese      & 6.00 & 19.40 & 7.49 & 7.90 & 7.00 & \underline{4.92} & \textbf{4.81} \\
        Code-Switching & 8.80 & 5.30 & 6.49 & 6.37 & 6.90 & \underline{4.35} & \textbf{4.20} \\
        \bottomrule
    \end{tabular}%
    }
\end{table}

\paragraph{Context utilization.}
Table~\ref{tab:context} evaluates context utilization on two complementary test sets. The intra-utterance context test set focuses on homophonic and phonetically similar target words whose recognition benefits from contextual cues within the same utterance, without externally supplied context. The dialogue-context test set covers question--answer interactions in real-world business scenarios and evaluates the use of preceding conversational turns. Hy ASR 3.0 achieves the highest target-keyword recall among the evaluated systems on both sets, reaching 61.88\% and 83.94\%, respectively. These results exceed the strongest non-Hy competitor, Seed 2.0 Lite, by 7.24 and 1.74 percentage points. They are cross-system comparisons rather than within-model ablations with and without context.

\begin{figure}[H]
    \centering
    \includegraphics[width=\linewidth]{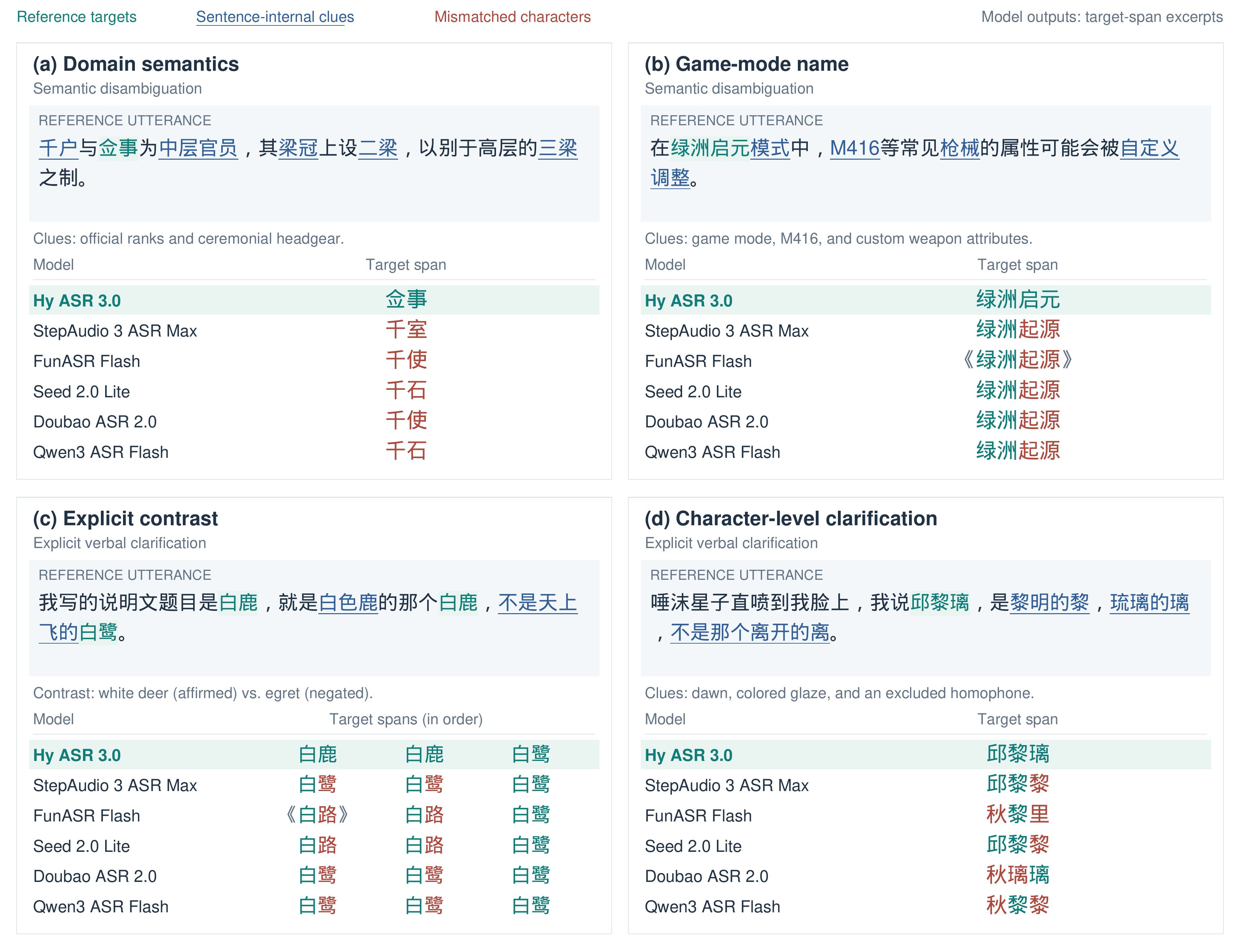}
    \caption{Examples of intra-utterance disambiguation.
Panels (a--b) illustrate semantic disambiguation, while panels
(c--d) illustrate explicit verbal clarification.
All contextual cues are contained within the spoken utterance.
Model outputs are shown as target-span excerpts.}
    \label{fig:intra-utterance-cases}
\end{figure}

Figure~\ref{fig:intra-utterance-cases} presents four selected examples illustrating semantic disambiguation and explicit verbal clarification, all evaluated without external context. In the first two, Hy ASR 3.0 correctly recognizes an uncommon historical office title and a game-mode name in their domain contexts, while the compared systems produce phonetically similar alternatives. The other examples distinguish an affirmed ``white deer'' from a negated ``egret'' and resolve two adjacent homophonic characters in a person's name using character-level explanations. StepAudio 3 ASR Max and Seed 2.0 Lite correctly transcribe the explanations in the latter example but still repeat the same character in the name, whereas Hy ASR 3.0 preserves the intended distinction. These cases illustrate outputs consistent with sentence-level semantic cues and explicit orthographic clarification. They are qualitative illustrations rather than controlled ablations of the contribution of context.

\begin{table}[H]
    \centering
    \small
    \caption{Context Utilization Evaluation (Recall, \%). \textbf{Bold} and \underline{underlined} values mark the best and second-best results in each row, respectively.}
    \label{tab:context}
    \resizebox{\textwidth}{!}{%
    \renewcommand{\arraystretch}{1.23}
    \begin{tabular}{lccccccc}
        \toprule
        \textbf{Testset} & \textbf{Qwen3 ASR Flash} & \textbf{FunASR Flash} & \textbf{Doubao ASR 2.0} & \textbf{Seed 2.0 Lite} & \textbf{StepAudio 3 ASR Max} & \textbf{Hy ASR 3.0 Preview} & \textbf{Hy ASR 3.0} \\
        \midrule
        Intra-utterance context & 46.60 & 51.16 & 51.86 & 54.64 & 50.75 & \underline{58.12} & \textbf{61.88} \\
        Dialogue QA context & 71.68 & 77.33 & 78.47 & \underline{82.20} & 69.32 & 79.44 & \textbf{83.94} \\
        \bottomrule
    \end{tabular}
    }
\end{table}

\paragraph{Robustness.}
Table~\ref{tab:robustness} evaluates recognition robustness on long-form speech, a combined child-and-older-adult speech set, and a combined evaluation of low-volume speech and speech in noisy environments. Hy ASR 3.0 achieves the lowest CER across all three sets, at 2.35\%, 6.25\%, and 7.49\%, respectively. Relative to the strongest non-Hy competitor in each setting, these correspond to CER reductions of 34.4\%, 2.8\%, and 19.2\%. The age-related benchmark combines children and older adults, so its result does not establish separate improvements for each age group.

\begin{table}[H]
    \centering
    \small
    \caption{ASR Robustness Evaluation (CER, \%). \textbf{Bold} and \underline{underlined} values mark the best and second-best results in each row, respectively.}
    \label{tab:robustness}
    \setlength{\tabcolsep}{3pt}
    \resizebox{\textwidth}{!}{%
    \renewcommand{\arraystretch}{1.23}
    \begin{tabular}{lccccccc}
        \toprule
        \textbf{Testset} & \textbf{Qwen3 ASR Flash} & \textbf{FunASR Flash} & \textbf{Doubao ASR 2.0} & \textbf{Seed 2.0 Lite} & \textbf{StepAudio 3 ASR Max} & \textbf{Hy ASR 3.0 Preview} & \textbf{Hy ASR 3.0} \\
        \midrule
        Long audio & 3.86 & 3.98 & 5.60 & 4.99 & 3.58 & \underline{2.96} & \textbf{2.35} \\
        Child \& Elderly & 9.98 & 7.87 & \underline{6.43} & 10.49 & 9.90 & 7.18 & \textbf{6.25} \\
        Low-volume \& Noisy & 14.72 & 9.27 & 15.75 & 17.31 & 43.02 & \underline{8.48} & \textbf{7.49} \\
        \bottomrule
    \end{tabular}
    }
\end{table}

\paragraph{Dialect recognition.}
Dialect recognition presents unique challenges because many dialects lack standardized writing systems or consistent orthographic conventions. This complicates data collection and annotation, model training, and evaluation, as the same spoken content may be transcribed in multiple valid ways, making conventional WER an unreliable measure of recognition quality. For dialects without established writing conventions, we therefore adopt a translation-based approach to represent the spoken content in a more consistent written form. During model development, we use an LLM-based judge to assess dialect transcription quality along two complementary dimensions: semantic error rate, which measures errors in conveyed meaning, and transcription style consistency, which evaluates the consistency of the written representation. These metrics provide practical signals for iterative model development, but may not fully capture dialect recognition quality. We therefore consider human evaluation the final assessment and conduct a comprehensive study covering 13 dialect categories and 91 subregional varieties, comparing Hy ASR 3.0 with FunASR Flash and Doubao ASR 2.0. These two systems are selected as baselines based on their strong performance in our objective evaluations, providing a comparison against competitive ASR systems. The results presented below are based on this final human evaluation.

We assess three complementary aspects of transcription
quality. \textit{Strict recognition rate} measures the percentage
of transcriptions judged entirely correct in their recognized
content. \textit{Lenient recognition rate} measures the percentage
judged semantically correct, allowing wording differences that
do not alter the meaning of the speech.
\textit{Formatting compliance rate} measures the percentage
judged free of formatting issues, including punctuation,
inverse text normalization (ITN) and fluency. These human-rated metrics
separately assess content correctness, semantic correctness,
and output formatting rather than character- or word-level
error rates.

Overall, Hy ASR 3.0 achieves a strict recognition rate of 64.7\%,
exceeding FunASR Flash and Doubao by 2.8 and 12.5 percentage
points, respectively. It also achieves a formatting
compliance rate of 80.3\%, outperforming the two systems by
11.4 and 13.7 percentage points, respectively. Under the lenient
recognition criterion, Hy ASR 3.0 outperforms Doubao but trails
FunASR Flash by 1.0 percentage point. These results demonstrate
stronger overall performance in fully correct transcription
and formatting compliance among the compared systems, while
leaving room for improvement in semantic correctness under
the lenient criterion.

\begin{figure}[H]
    \centering
    \includegraphics[width=\linewidth]{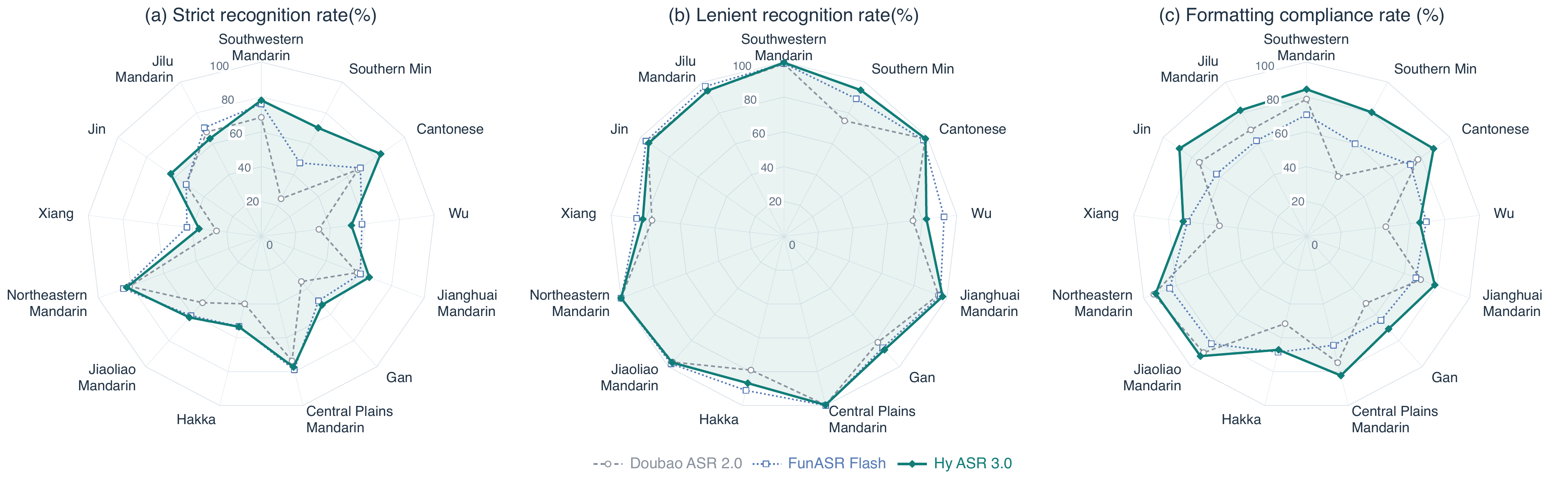}
    \caption{Human evaluation of dialect recognition.}
    \label{fig:dialect-three-metrics}
\end{figure}

Figure~\ref{fig:dialect-three-metrics} presents the per-category
results for the 13 dialect categories shown. Hy ASR 3.0 achieves
particularly strong strict recognition results on Southern Min
and Cantonese, reaching 70.3\% and 83.3\%, respectively, exceeding
the strongest competitor in each category by 22.7 and
14.2 percentage points. Nevertheless, performance varies across
categories: FunASR Flash remains stronger on Wu under all three
metrics. The category-level results thus complement the overall
scores by highlighting both the model's strengths and its
remaining gaps.

Overall, Hy ASR 3.0 outperforms FunASR Flash and Doubao ASR 2.0 in dialect recognition under the strict criterion. Under the lenient criterion, it achieves performance comparable to FunASR Flash while maintaining an advantage over Doubao ASR 2.0. Furthermore, Hy ASR 3.0 delivers substantially higher text quality compliance, reflecting its ability to produce accurate, fluent, and well-formatted transcriptions.

\section{Conclusion}
\label{sec:conclusion}
We presented Hy ASR 3.0, an LLM-based ASR system combining strong recognition accuracy, flexible use of contextual information, and robustness across diverse speech conditions. The system integrates an acoustic encoder, an audio adapter, and an LLM decoder, supported by large-scale pre-training and multi-stage post-training. Evaluations on public and internal benchmarks demonstrate leading aggregate recognition performance for Mandarin Chinese, English, and Cantonese among the evaluated systems. Human evaluation further highlights its strengths in strict dialect recognition and transcription formatting, with broadly comparable lenient recognition to FunASR Flash. Beyond standard transcription, Hy ASR 3.0 achieves strong results on evaluations of external dialogue context and intra-utterance disambiguation, as well as long-form speech, a combined child-and-older-adult speech set, and a combined evaluation of low-volume speech and speech in noisy environments. Remaining gaps on particular dialect categories and accented speech motivate further work on more consistent recognition across regional and acoustic variation, together with more detailed evaluation of contextual inputs and the contributions of individual training stages.

\section{Authors}

\subsection*{Core Contributors\textsuperscript{1}}

{\linespread{1.2}\selectfont
\noindent
Aoxiong~Yin, Auden~Li\smash{\textsuperscript{$\dagger$}}, Chao~Huang, Chuanyang~Hu, Eitan~Wang\smash{\textsuperscript{$\dagger$}}, Ethan~Zou, Fangming~Feng, Guangyu~Chen, Guinan~Li, He~Wang, Jiahong~Li, Meng~Yu, Mingyu~Cui, Peng~Wang, Rongjie~Li, Steve~Yves\smash{\textsuperscript{$\dagger$}}, Teddy~Sun, Tian~Zheng, Wei~Liu, Yang~Zhou, Yixuan~Zhang, Zhennan~Lin, Zhihang~Xu, Zhuohao~Li, Ziqiang~Zhang.
\par}

\vspace{1em}

\subsection*{Contributors\textsuperscript{1}}

{\linespread{1.2}\selectfont
\noindent
Anji~Zhang, Bang~Wang, Caisheng~Ouyang, Chengqi~Zhu, Chong~Li, Conglian~Xie, Dongdong~Wang, Edward~Wang, Evelynn~Liu, Ferman~Yang, Feng~Lu, Gang~Zhang, Jathon~Zhang, Jaxl~Liu, Jiachen~Luo, Jian~Li, Jingxi~Gui, Kysun~Wang, Liang~Dong, Longsha~Zhou, Na~Hu, Qi~Chen, Rilin~Chen, Rui~Yuan, Shanshan~Sang, Siyu~Yan, Steven~Hong, Taihui~Wang, Weiwei~Li, Xiangyu~Zhang, Xiaotong~Yang, Xue~Hu, Xue~Jiang, Yameng~Li, Yifu~Sun, Yuhuan~Zhou, Zhiang~Chen, Zhibin~Wang.
\par}

\begingroup
\renewcommand\thefootnote{$\dagger$}\footnotetext{Project Leader (\texttt{audenli@tencent.com}, \texttt{ethwang@tencent.com}, \texttt{steveyves@global.tencent.com}).}
\endgroup
\footnotetext[1]{Alphabetical order by first name.}

\clearpage
\bibliographystyle{colm2024_conference}
\bibliography{references}

\end{document}